\documentclass{article}
\usepackage{arxiv}

\usepackage[utf8]{inputenc} 
\usepackage[T1]{fontenc}    
\usepackage{hyperref}       
\usepackage{url}            
\usepackage{booktabs}       
\usepackage{amsfonts}       
\usepackage{nicefrac}       
\usepackage{microtype}      
\usepackage{cleveref}       
\usepackage{lipsum}         
\usepackage{graphicx}
\usepackage{natbib}
\usepackage{doi}

\usepackage{subcaption}
\usepackage{scalerel}
\usepackage{siunitx}

\usepackage{setspace}
\title{
  Social-spatial dependencies for learning visual navigation
}

\date{\today}

\usepackage{authblk}

\author[1]{
	{Patrick Govoni\thanks{Corresponding Author E-mail: \texttt{pgovoni21@gmail.com}}}
}
\author[1,2,3]{
	{Pawel Romanczuk}
}
\affil[1]{Institute for Theoretical Biology, Department of Biology, Humboldt Universität zu Berlin, Berlin, Germany}
\affil[2]{Science of Intelligence, Research Cluster of Excellence, Berlin, Germany}
\affil[3]{Bernstein Center for Computational Neuroscience, Berlin, Germany}

\renewcommand{\shorttitle}{}

\hypersetup{
pdftitle={Social-spatial dependencies and interfaces for learning visual navigation},
pdfsubject={},
pdfauthor={Patrick Govoni, Pawel Romanczuk},
pdfkeywords={},
}

\begin{document}
\maketitle

\begin{abstract}

Navigation for social organisms rarely is a fully independent activity.
Group structure and dynamics, as well as embodied interactions, critically influence useful behavior. 
Individual neural network controlled agents are trained to navigate in different social contexts, 
where social dependence and behavioral strategy learned is determined by relative task performance and spatial effect. 
Increasing high quality social information drives phase transitions from individual to following navigational strategy, 
and to collision avoidance in response to a crowded foraging patch.
Predictable, nonstationary environmental dynamics drive behavioral hybridization between individual and social navigation, far and near the patch.
Our findings challenge the approach of only inspecting individual behavior for social organisms
and highlight the importance of taking a bottom-up approach in understanding how organisms behave.

\end{abstract}




\section{Results}

\subsection*{Model design}

\begin{figure}[htb]
	\centering
	\includegraphics[width=1\linewidth,trim={0 0 0 0},clip]{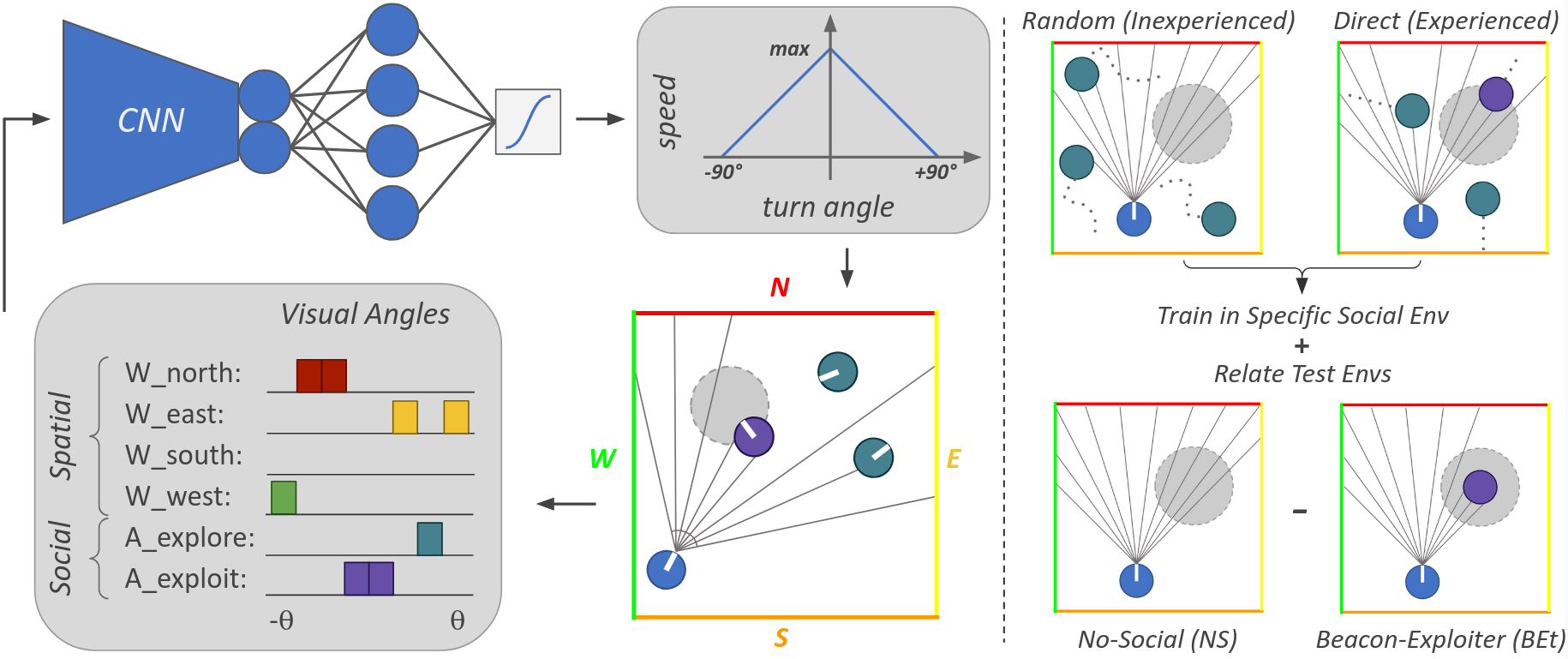} 
	\caption{
		\textbf{Agent flow \& train-test methodology.}
		Left: (clockwise from bottom left) visual encoding, information processing, action conversion, environment update.
		Visual encoding identifies walls and other agents corresponding to retinal angles of a raycast 
		(between -$\theta$ \& $\theta$ field of view limits).
		Visual information for the trained agent (blue) passes through 
		convolutional neural network, perceptron, linear output layer, and hyperbolic tangent tranformations 
		to directly represent both turning angle and speed via a linear function, 
		which updates agent position and orientation for the next timestep.
		Untrained agents can be identified by their action-location status: on the patch and exploiting (purple), or off the patch and exploring (cyan).
		Right: training with varying group structure and testing with a point perturbation;
		(top) two example environments with low/high social skill,
		where dotted lines showing potential paths of the others;
		(bottom) measuring learned social dependency by comparing behavior in perturbed test environments.
		}
	\label{fig:intro}
  \end{figure}

The task is to navigate to a hidden patch located at a fixed position in a minimal, square environment
(Fig. \ref{fig:intro}, left).
Four walls can be distinctly identified, where the four corners comprise the salient landmarks.
The agents, both trained and untrained, are initialized randomly in the environment for each simulation.
Additionally, two types of untrained agents provide the option of social navigation with differing skill levels:
experienced agents that walk directly to the patch and inexperienced agents that walk randomly about the environment.
Population-level skill level and density were varied as the relative proportion and total number of direct and random agents in the training environment.

The trained agent visually perceives the environment and other agents by raycast,
inspired by~\cite{bastienModelCollectiveBehavior2020}
and extending the non-social navigation in~\cite{govoniVisuospatialNavigationDistance2024}.
Eight rays extend from a center retina to the first collided object: boundary wall or another agent.
Collisions one-hot encode which objects are at which egocentric angles.
Other agents can be distinguished according to action-location status: whether the agent is on the patch and exploiting, or off the patch and exploring.

Body collisions, both agent-agent and agent-wall, are simulated with absorbing boundary conditions, where only movement directions in non-colliding directions are permitted.
To break stalemates for direct agents, a random direction is held for five time units.
The total number of untrained agents has been kept between zero and five in order to reduce issues at the patch boundary
(Fig. \ref{fig:num-limit}).

Agent networks combines a convolutional neural network (CNN), single perceptron layer, and linear output layer.
Visual input compresses into a single action output that updates agent location and orientation for the next timestep,
constrained as a linear ratio between turning angle and speed, describing the need to slow down in order to turn.

Performance is calculated as the time taken to reach the patch, 
plus the remaining distance if the agent has not reached the patch within the simulation time limit.
While the first term directly represents navigation ability, the second term guides initial learning behavior.
An evolutionary strategies (ES) algorithm was used for optimization, 
its implicitly explorative population-based approach was chosen over single-agent reinforcement learning (RL).


\subsection*{Task dependence}

Training agents in this environment affords at least two potential learned behaviors:
an individual navigation strategy that bases movement decisions relative to spatially fixed landmarks,
or a social strategy consisting of following (or avoiding) the other, untrained agents.
As illustrated in Fig. \ref{fig:intro} (right-top),
the untrained agents either: walk directly towards the patch, or pursue random walks without regards to patch location.
The ratio between direct and random agents reflects the level of observable social skill (or navigational experience) in the environment,
while the total number of untrained agents reflects social density.
This social skill ratio was hypothesized to negotiate the degree to which trained agents learn to depend on social information.
While it is possible to learn an individual spatial navigation strategy, 
as previously demonstrated in non-social environments \cite{govoniVisuospatialNavigationDistance2024},
a higher ratio of direct to random agents, thus higher quality social information, was assumed to favor social navigation.

After training agents in various environments of specific social skill and density, 
the aim is to distinguish dependency on social information from that on spatial landmarks.
Trained agents were tested in two separate environments
(Fig. \ref{fig:intro}, right-bottom):
one without any other agents (non-social, NS), and one with an exploiting agent on the patch center (beacon-exploiter, BEt).
In this format, learned navigational strategy was tested with a point perturbation 
without influence from social skill, density, or spatiotemporal dynamics from untrained agents.
Relating task performance (average travel time to reach the patch) between these two test environments establishes a measure for social dependence.
Agents with similar travel times across both environments do not depend on social information for navigation.
Positive travel time difference indicates beneficial social dependency, 
with the value indicating extra time the agent would need to reach the patch on its own rather than with a beacon-exploiter, if it can.
Negative travel time difference indicates social distraction, a maladaptive social dependency.

\begin{figure}[!h]
	\centering
	\includegraphics[width=.94\linewidth,trim={0 0 0 0},clip]{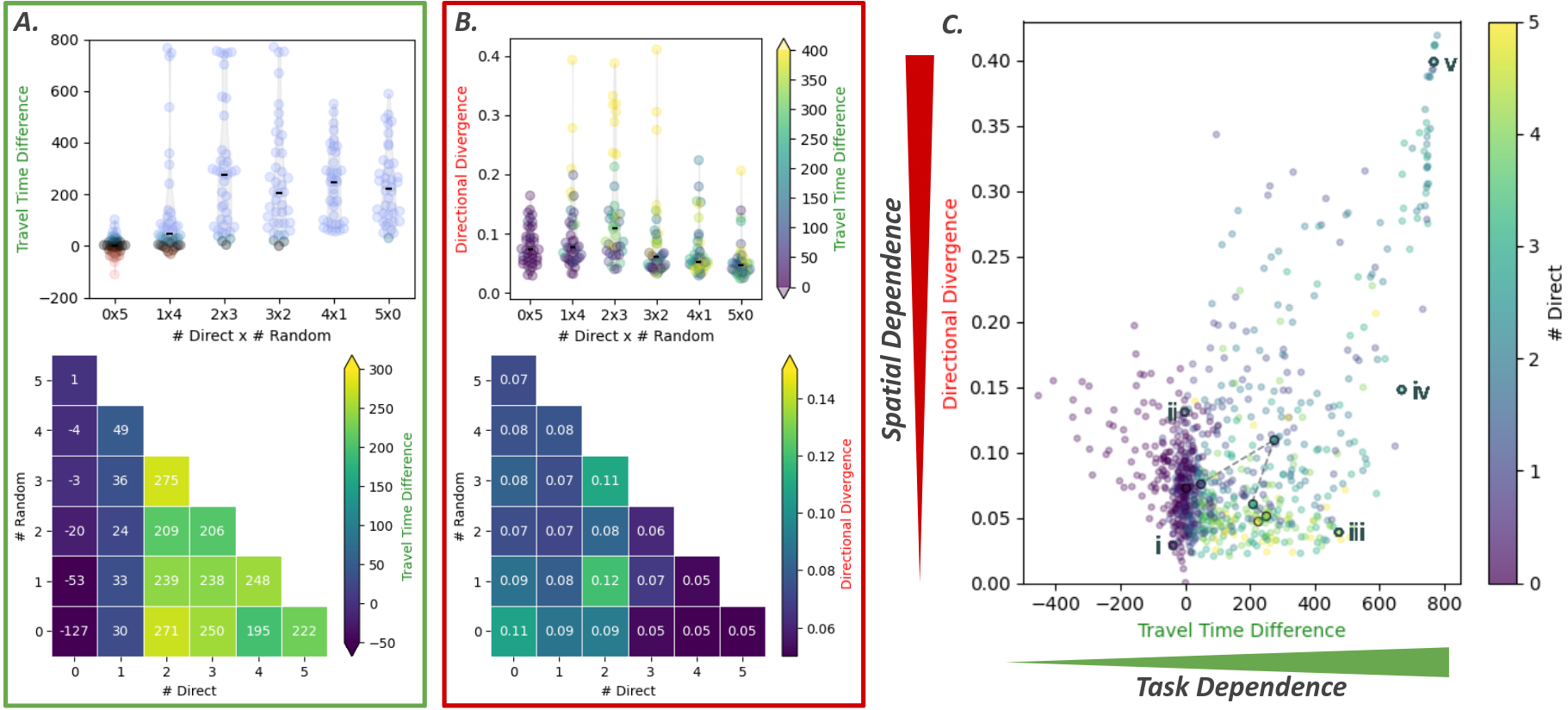} 
	\includegraphics[width=.94\linewidth,trim={0 0 0 0},clip]{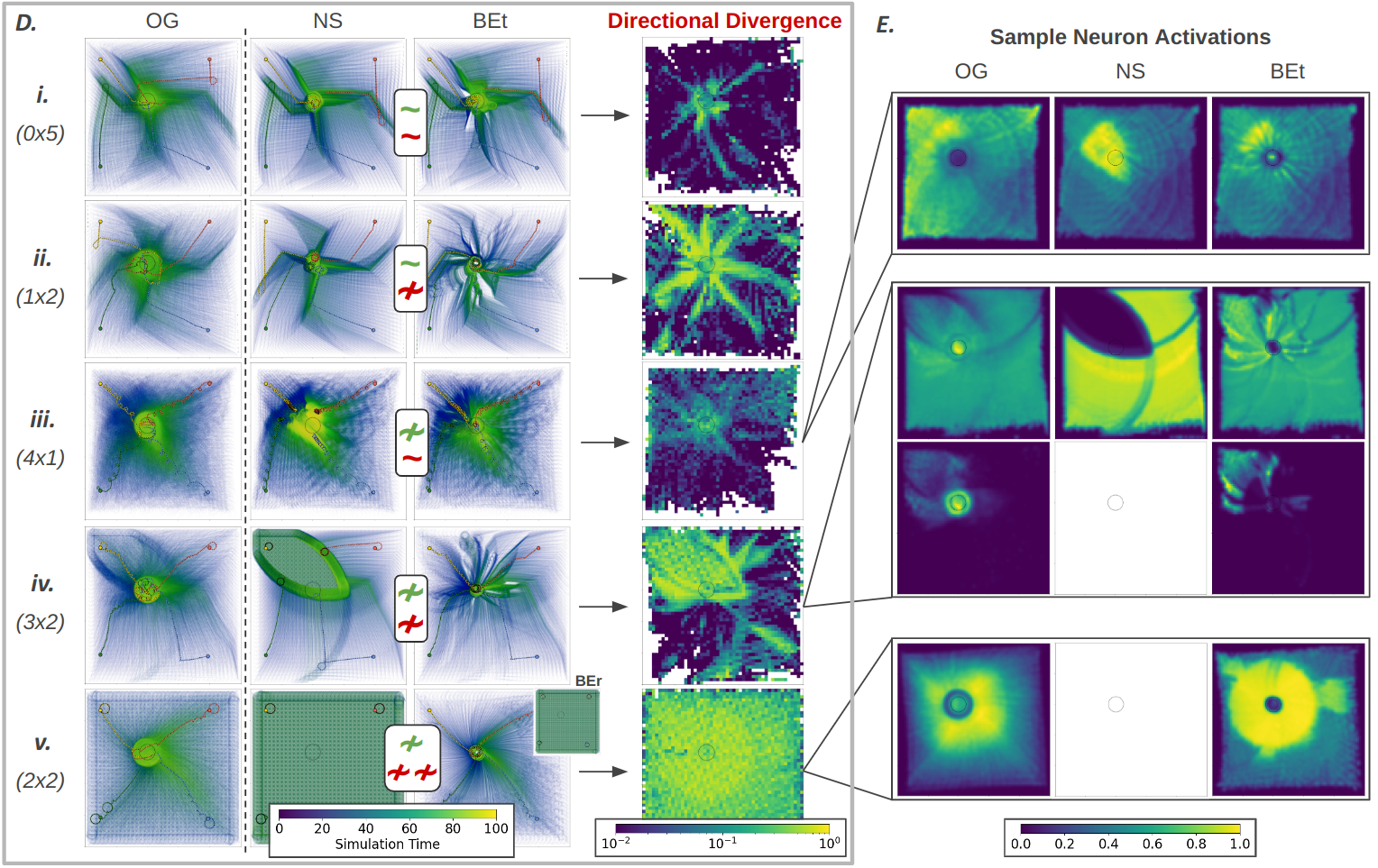}
	\caption{
		\textbf{Learned task \& spatial dependencies, movement \& neural activation maps.}
		A: 
		relative task performance between non-social (NS) \& beacon-exploiter (BEt) test environments,
		as scatter plot for N=5 (top) \& stair plot for N<=5 (bottom), 
		where N is total number of untrained agents,
		varying the ratio between direct/random untrained agents per condition.
		All data (40 runs per condition, 5000 initializations averaged per test) aggregated with medians as scatter plot horizontal lines and stair plot color \& number.
		B: 
		directional divergence compares spatial-orientation distributions between NS/BEt test environments (plotted here as the spatial average).
		C: 
		data from A/B as 2D plot, 
		where numerals (i-v) correspond to examples in D,
		and black outlined/connected points as population-level medians for the N=5 conditions in A/B scatter plots.
		D: 
		Learned behavior of five example agents plotted as movement trajectory maps,
		in the original training (OG), NS, BEt (inset: beacon-explorer (BEr)) environments.
		Five agents (i-v) trained in (n x m) social environments,
		with n direct \& m random untrained agents.
		Green tilde indicate (dis)similarity in task performance, red tilde indicate spatial (dis)similarity.
		Directional divergence maps outline spatial characteristics of the measure in B.
		E:
		Neuron activations illustrate varying spatial-social behavior:
		agent iii/iv - individual-social navigation hybridization;
		agent v - social action-status discrimination.
		}
	\label{fig:dep-dirdiv}
  \end{figure}

Keeping social density constant at five untrained agents and varying the skill ratio, 
we find that indeed, more direct agents result in a greater learned social dependence
(Fig. \ref{fig:dep-dirdiv}A, top).
At the population level, the relation takes the form of a step function:
while agents can learn to be socially dependent with only one direct agent, the median shifts up at two and remains at that level even for an all-direct environment.
Relaxing the constraint on social density, we find that this step threshold is true regardless of the number of random agents in the environment
(Fig. \ref{fig:dep-dirdiv}A, bottom).
The environment needs at least two direct agents for an agent to learn a navigational dependency on social information,
and more direct agents do not increase this dependency.


\subsection*{Spatial dependence}

Navigational performance alone, however, was found to not capture the degree to which social information is used spatially across the environment.
Those with low travel time difference can differ markedly in how their trajectories are affected by other agents 
(Fig. \ref{fig:dep-dirdiv}D, i-ii),
likewise relying on others to reach the patch may not correlate with spatial dependence
(Fig. \ref{fig:dep-dirdiv}D, iii-v).
Building on an entropic measure from previous work~\cite{govoniVisuospatialNavigationDistance2024},
directional divergence uses Jensen-Shannon divergence on spatially binned distributions of agent orientations.
For a given spatial bin, the collection of agent orientations (from intersecting movement trajectories)
can either be similar across non-social and beacon-exploiter test conditions (i/iii), or widely vary (ii/iv-v).
Similar directional distributions result in low divergence, corresponding to a small social impact on trajectories for that location,
while varying directional distributions or high divergence correspond to a large spatial impact.
Thus beyond mean-field parameterization for this metric, directional divergence can be spatially uniform (v) or heterogeneous (ii/iv).

From zero to two direct agents, directional divergence tends to increase along with travel time difference
(Fig. \ref{fig:dep-dirdiv}B/C/D).
Agent behavior varies along this axis from individual navigation with little social interaction (i-ii), to fully dependent social following (v). 
The measures decouple with greater than three direct agents, with directional dependence decreasing to lower median levels than for all-random environments.
Low spatial dependence in high direct environments contradicts our expectation:
with high quality social information, learned strategies depend on social information to complete the task, 
yet their spatial trajectories are not substantially affected.
For much of the spatial landscape these agents navigate on their own via landmarks,
yet critically depend on others to finally reach the patch 
(Fig. \ref{fig:dep-dirdiv}D, iii-iv),
in other words, hybridizing individual and social navigation.

Activation maps of sample neurons portray elements of these spatial phenomena
(Fig. \ref{fig:dep-dirdiv}E).
For the two hybrid examples (iii-iv), NS activation profiles clearly demarcate boundaries for individual navigation outside the patch,
sharply contrasting with smoother transitions in social conditions (OG/BEt).
In the periphery, agents move along predictable paths regardless of social presence (lower local directional divergence),
whereas movement is significantly impaired or stopped at these boundaries in the non-social context.
The boundaries thus govern an interface where the agent must switch from individual to social navigation in order to complete its route.
In addition to these interface neurons, others that activate only with social input provide support closer to the patch
(Fig. \ref{fig:dep-dirdiv}E, 3rd row).
Pure followers contain neurons with similar capacities, where activation only reacts to social visual observations (v),
while also demonstrating spatial variability with respect to distance from the observed agent.


\begin{figure}[htb]
	\centering
	\includegraphics[width=.8\linewidth,trim={0 0 0 0},clip]{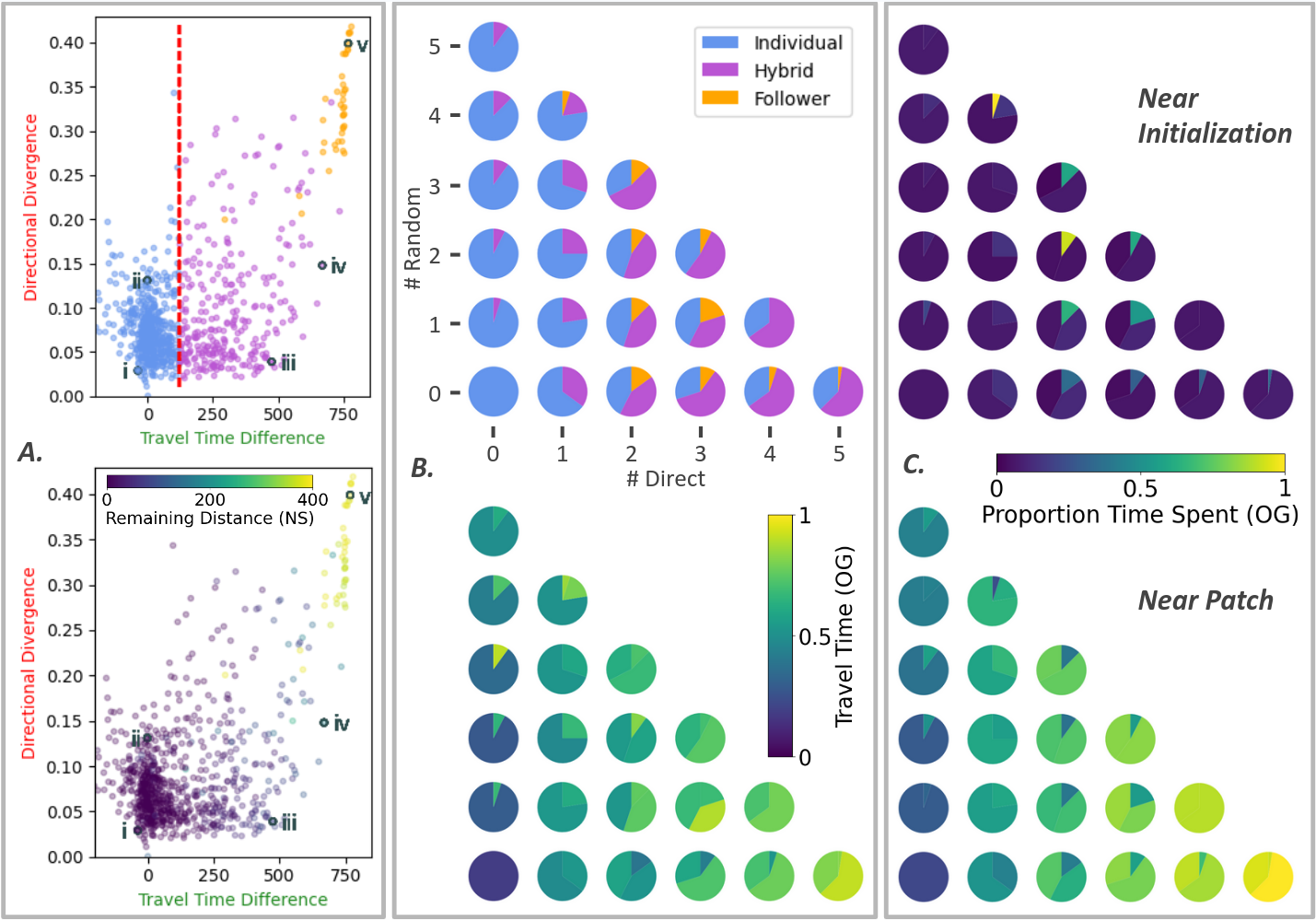} 
	\caption{
		\textbf{Categorizing learned behaviors.}
		A: 
		2D plots replicating Fig. \ref{fig:dep-dirdiv}C,
		with the top separating agents into 3 categories:
		with navigators defined as less than 120 travel time difference (red dashed line),
		followers as above 400 remaining distance (shown below),
		and hybrids as those not fulfilling either criteria.
		B:
		pie charts organized into stair plots separated by training environment 
		(top: categories, bototm: travel time in the original trained environment), 
		as in Fig. \ref{fig:dep-dirdiv}A/B bottom.
		C:
		trajectory data as used for Fig. \ref{fig:dep-dirdiv}D (OG)
		with time split among (near-initialization, in-transit, near-patch, at-patch) categories,
		both categories shown here calculated within 50 spatial units of either.
		}
	\label{fig:types}
  \end{figure}

\subsection*{Behavioral categorization}

While two behaviors were expected to be learned by the networks, individual navigation or social following, a third strategy merges the two.
Quantifying the relative distribution between the three categories is considered in two parts.
Individual navigation (with no social dependence) is characterized by low travel time difference
(Fig. \ref{fig:types}A, top, red dotted line).
Following is defined as agents that do not move significantly from their initial locations without social input,
i.e. high remaining distance in the NS test condition
(Fig. \ref{fig:types}A, bottom).
Both thresholds are drawn at histogram boundaries (Fig. \ref{fig:histo-thresh}).
Followers tend towards high travel time difference and directional divergence, but not necessarily. 
Likewise, agents with low remaining distance does not fully align with individual navigation.

The share in following as well as hybrid strategies increases from zero to two direct agents
(Fig. \ref{fig:types}B, top).
From three to five, the share of followers drops, absorbed by the share of hybrids, 
while the individual navigators keep steady at about 1/3.
The performance of each category
(Fig. \ref{fig:types}B, bottom)
shows this persistence individual navigation efficiency with the exception for 2-4 direct agent environments with no random agents,
as well as a general decrease in performance as the number of direct agents increase.

These categorical trends can be explained by integrating over route trajectories.
Using the same data plotted in Fig. \ref{fig:dep-dirdiv}D for the original (OG) training environment,
trajectories were temporally integrated according to four criteria:
near-initialization (<50 units), in-transit, near-patch (<50 units), and at-patch (finished)
(Fig. \ref{fig:freqs-supp}).
The first grouping (Fig. \ref{fig:types}C, top)
shows how only followers wait at the start for any significant period of time, and those trained without random agents wait less.
Coupling with analysis relating travel times in BEt and beacon-explorer (BEr) conditions
(Fig. \ref{fig:ETER}),
this can be explained as followers in mixed direct-random environments learn to wait for exploiting agents, as random agents can lead astray.
While that followers tend to be faster in-transit
(Fig. \ref{fig:freqs-supp}),
the longer waiting time at the start can even lead to lower overall performance.
Thus, hybridizing individual navigation at the periphery and using social knowledge near the patch
takes advantage of the predictable environment transition delay for direct agents to reach the patch
(Fig. \ref{fig:first-passage-time})

\begin{figure}[htb]
	\centering
	\includegraphics[width=.72\linewidth,trim={0 0 0 0},clip]{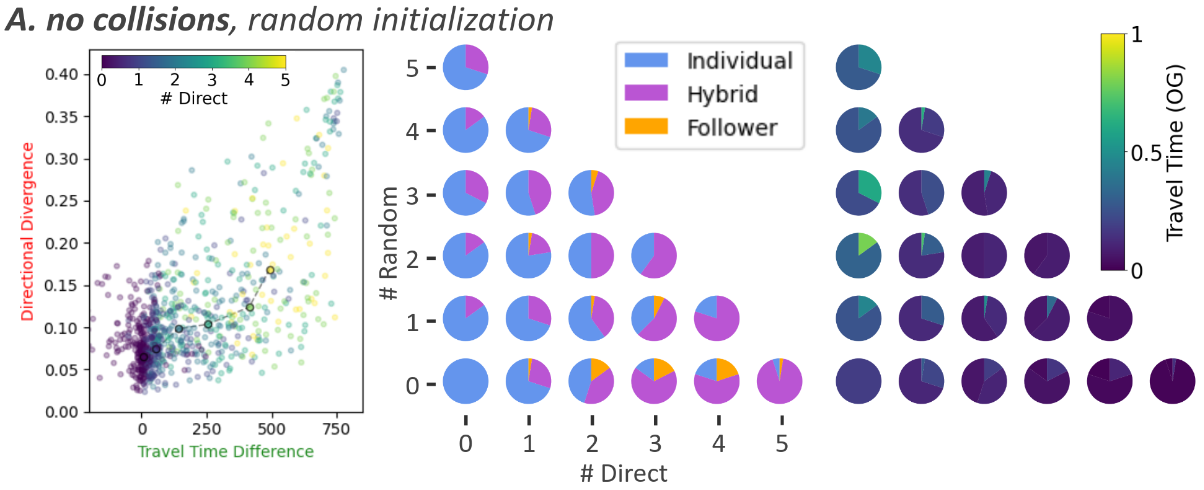}
	\includegraphics[width=.72\linewidth,trim={0 0 0 0},clip]{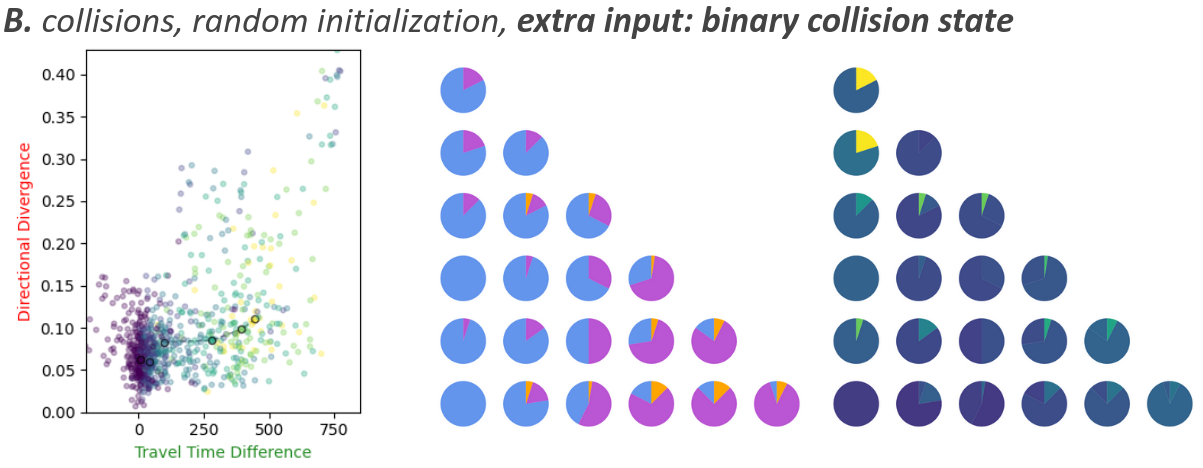}
	\includegraphics[width=.72\linewidth,trim={0 0 0 0},clip]{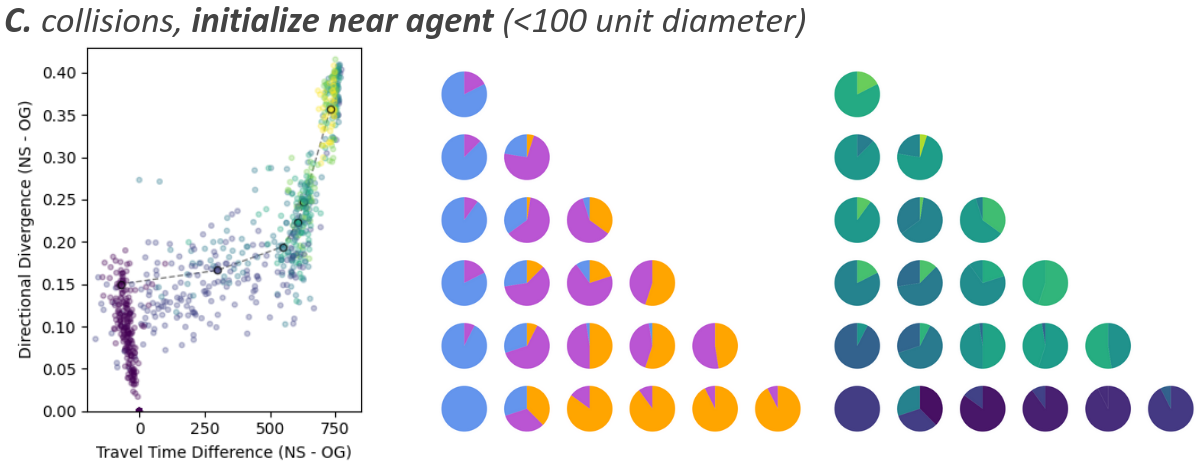}
	\includegraphics[width=.72\linewidth,trim={0 0 0 0},clip]{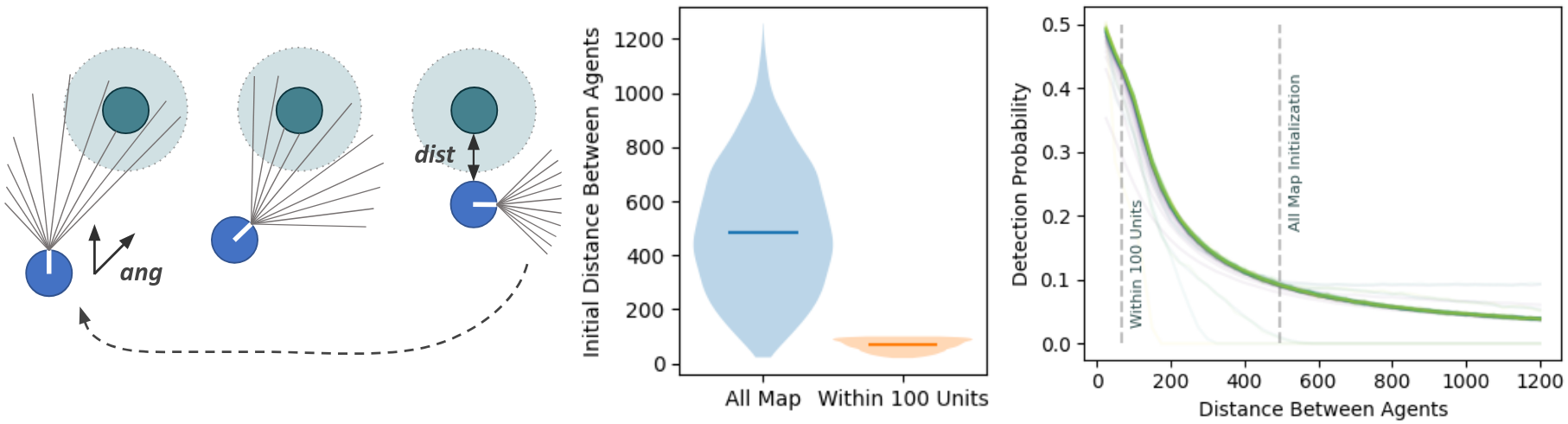}
	\caption{
		\textbf{Model Extensions: Embodiment/Perceptibility.}
		Results rerun with 3 changes to simulation parameters.
		Plots follow Fig. \ref{fig:dep-dirdiv}C/\ref{fig:types}B.
		A: 
		no collisions between agents.
		B: 
		focal agent receives extra binary input indicating current collision status to the linear output layer.
		C:
		untrained agents initialize within 100 spatial units of focal agent,
		where the test perturbations is NS-OG rather than NS-BEt (Fig. \ref{fig:intro}right-bottom) 
		given the unique initialization procedure. 
		}
	\label{fig:modelext}
  \end{figure}

The time spent near, but not on, the patch increases with the number of direct agents for individual navigators and hybrids
(Fig. \ref{fig:types}C, bottom),
suggesting that direct agents, arriving before the focal agent, block easy passage to the patch
(as seen with higher direct agent densities, Fig. \ref{fig:num-limit}).
This trend mirrors the decrease in the directedness of individual movement
(Fig. \ref{fig:dirent}),
indicating a shift from straight to looping movement styles
\cite{govoniVisuospatialNavigationDistance2024}.
Although looping decreases movement efficiency
(Fig. \ref{fig:freqs-supp}A, left),
it may afford the agent more reliable means to dodge other agents blocking the patch.


\subsection*{Extending the Model}

In order to perturb patch blocking, which favors collision avoidance looping over direct following movement,
three simulation changes were explored.
First, eliminating agent-agent collisions aligns increasing task with spatial dependencies
(Fig. \ref{fig:dep-dirdiv}C/\ref{fig:modelext}A, left).
Directional divergence no longer decorrelates from travel time difference at higher numbers of direct agents.
Allowing agents to pass through others physically eliminates the time cost of patch blocking
(Fig. \ref{fig:modelext-extra}A, right),
increasing fitness for the focal agents
(Fig. \ref{fig:modelext}A, right)
however this does not translate to agents more frequently learning to follow
(Fig. \ref{fig:modelext}A, center).
Rather, followers in mixed environments, with one or more random agents, are significantly less common than for simulations with collisions.

Second, collisions were allowed but the focal agent could sense its immediate collision state, in addition to visual input
(Fig. \ref{fig:modelext}B).
Though this model changes similarly does not show the social dependency decorrelation, directional divergence is lower than for collision-free simulations.
There is patch blocking, but less than the original simulation environment, leading to better relative fitness.
However as with above, a following strategy is less frequently learned in mixed environments.

The final simulation change was to restrict all social agents to initialize close to the focal agent
(Fig. \ref{fig:modelext}C).
As this concentration significantly affects the visual input reguarly experienced by the agent,
the dependency perturbations comparing a single beacon exploiter on the patch to a non-social test environment are less relevant to high total agent training conditions.
Though comparing non-social to the original training environment results in a density-dependent increase in directional divergence
(Fig. \ref{fig:dependencies-flip}),
task dependence and the N=5 decorrelation with higher direct agent environments show similar patterns as for metrics comparing NS/BEt.
For the initialize-close simluations, this reveals a strong nonlinear increase in both metrics 
as well as significantly greater follower frequencies relative to other simulation conditions. 
Mixed environments give rise to noticeably lower fitness, from longer waiting times at the start
(Fig. \ref{fig:modelext-extra}C, 3rd from left).
Patch blocking does not noticeably affect these agents since the direct agents, being spatially concentrated at simulation start, 
remain concentrated to one side of the patch at the end of their trajectories. 

Learning to follow other agents as a navigational strategy requires before movement.
A higher concentration of social agents nearby raises detection probability
(Fig. \ref{fig:modelext-extra}D),
while agents initialized randomly across the environment, at about five times the distance as the initialize-close simulations,
can be detected with roughly a quarter of the probability.
While these agents trained in high direct environments do not respond strongly to single beacons, 
they do to five agents on the patch, 
as well as when testing directional divergence within a closer, perceptible neighborhood of the patch. 

\section{Discussion}

Our study involved training a minimal perception-action agent to navigate to a hidden patch by visually perceiving boundary walls and other agents
(Fig. \ref{fig:intro}, left).
Each agent was simulated with a physical body, both occluding visual rays to walls and impeding trajectories via agent-agent collisions.

After training the neural network to behave within particular social environments, varying both skill and density,
learned behavior was tested via a point perturbation -- relating metrics in non-social (NS) and beacon-exploiter (BEt) environments
(Fig. \ref{fig:intro}, right).
Social dependence with respect to task performance (travel time difference) increase with the number of direct agents in the training environment,
though spatial dependence (directional divergences) decorrelates at high direct agent numbers
(Fig. \ref{fig:dep-dirdiv}A-C).

A spectrum of navigational styles encompass the metric space within the two metrics:
individual navigation via walls, social navigation via the other agents in the environment,
and hybrid variations in between.
Spatial interfaces demarcate behavioral changes between individual and social navigational movement,
as well as along a boundary with respect to following distance from another agent
(Fig. \ref{fig:dep-dirdiv}D-E).

Categorizing learned behaviors illustrate a decrease in follower counts and increase in hybrids in higher direct training environments
(Fig. \ref{fig:types}).
Those trained in mixed environments are more likely to develop a wait-and-see strategy, 
with capability of discerning between exploring and exploiting, in order to only follow agents once they reach the patch.
High numbers of direct agents tend to block the focal agent from accessing the patch, 
leading to shift toward looping movement patterns learned as means to avoid collisions and also describing the decrease in spatial dependence.

Three model extensions were explored to shift conditions causing patch blocking
(Fig. \ref{fig:modelext}).
Allowing agents to pass through each other and allowing agents to sense their collision state both result in correlation between the dependency metrics,
although following in mixed environments under these conditions is a less commonly learned strategy.
Initializing other agents close to the focal agent greatly increases both task and spatial dependence as well as boosts follower frequency.
This phenomenon is likely caused by a greater ability to visually perceive nearby agents.

\bibliographystyle{unsrt}
\bibliography{social}

\section{Supporting information}

\setcounter{figure}{0}
\renewcommand{\figurename}{Fig}
\renewcommand{\thefigure}{S\arabic{figure}}

\begin{figure}[htb]
	\centering
	\includegraphics[width=1\linewidth,trim={0 0 0 0},clip]{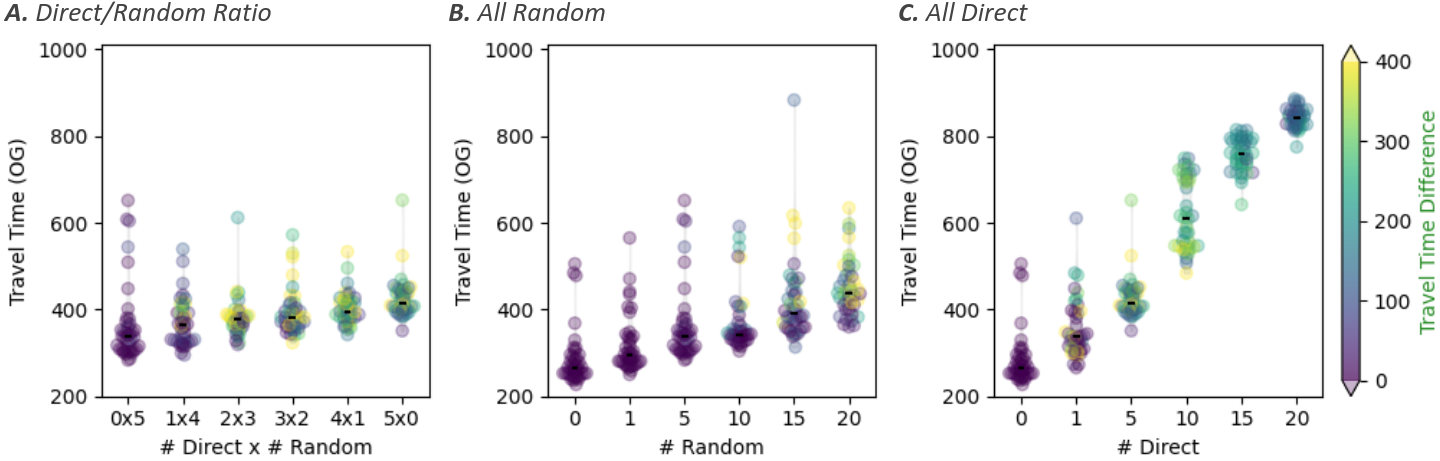}
	\caption{
		\textbf{Agent Number Limit.}
		Agent fitness as (travel) time taken to reach the patch in the original (OG) training environments,
		colored by the travel time difference, comparing non-social (NS) and beacon-exploiter (BEt) environments.
		The total number of agents used for our simulations was capped at five,
		in order to keep the travel time within 500 timesteps for both full random and full direct environments.
		At 10-20 numbers of direct agents (C), travel time increases significantly,
		while high random environments (B) does not affect travel time as much.
		}
	\label{fig:num-limit}
  \end{figure}

\begin{figure}[htb]
	\centering
	\includegraphics[width=.8\linewidth,trim={0 0 0 0},clip]{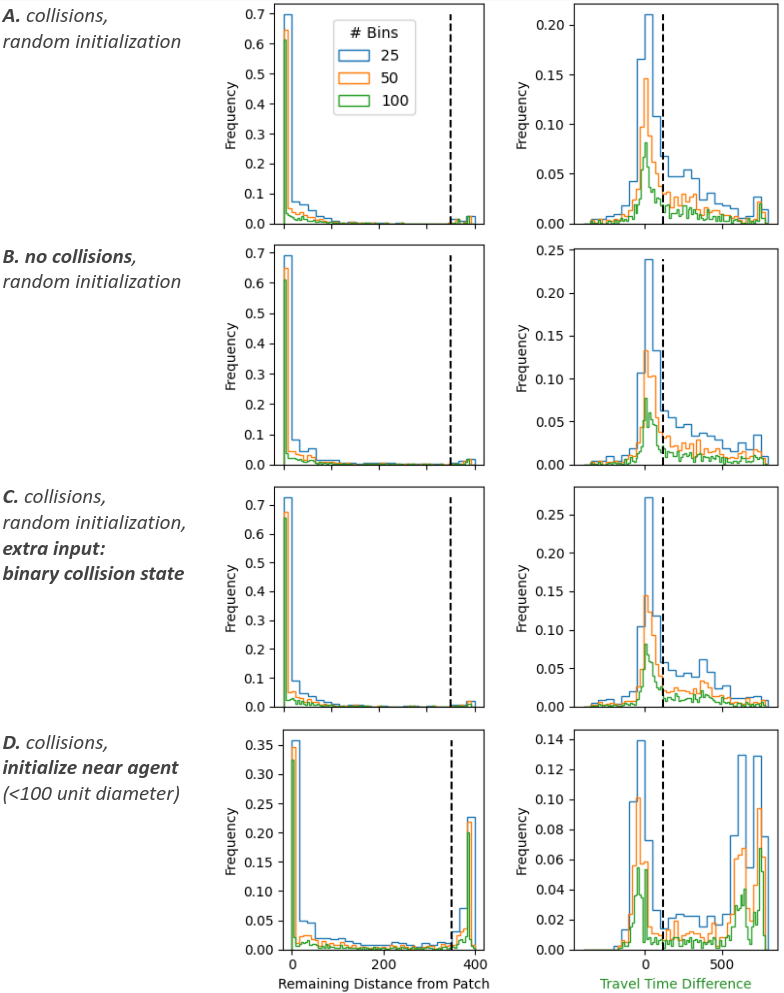}
	\caption{
		\textbf{Classification Histogram Thresholds.}
		All agent runs for the original simulation (A) and for the three model extensions (B-D)
		}
	\label{fig:histo-thresh}
  \end{figure}

\begin{figure}[htb]
	\centering
	\includegraphics[width=1\linewidth,trim={0 0 0 0},clip]{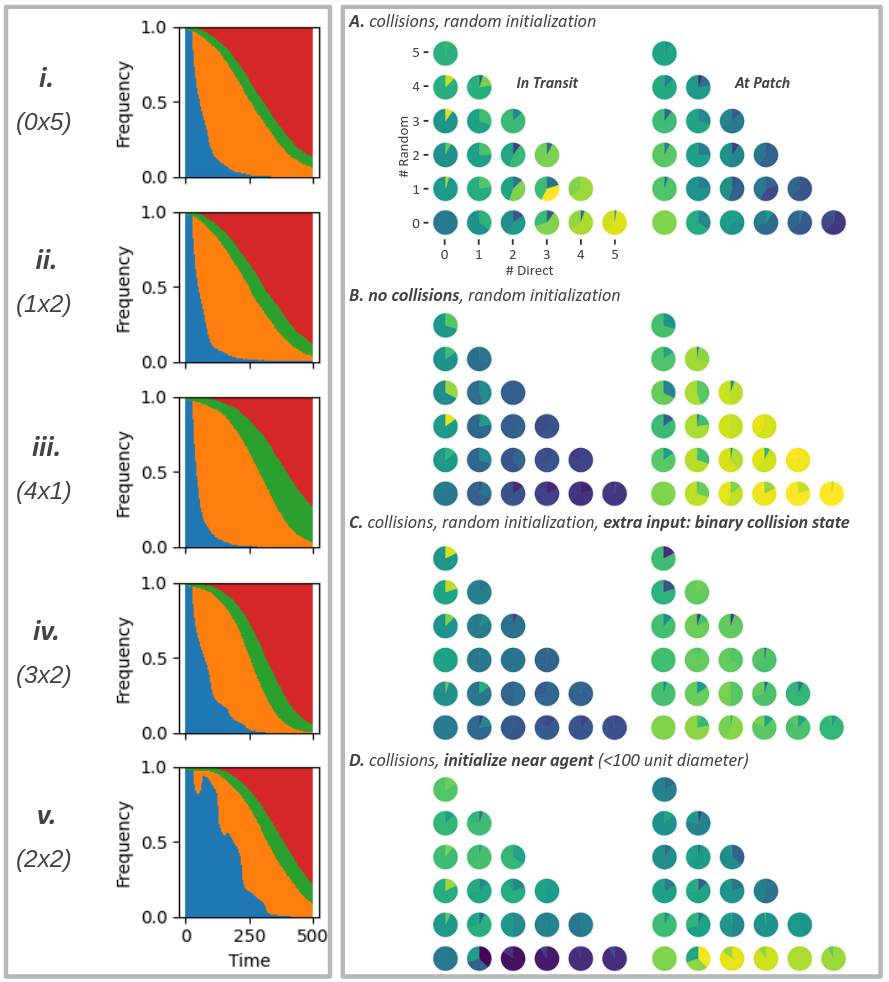}
	\caption{
		\textbf{Trajectory Integration: Extra.}
		}
	\label{fig:freqs-supp}
  \end{figure}

\begin{figure}[tb]
	\centering
	\begin{subfigure}[b]{\linewidth}
	  \centering
	  \includegraphics[width=1\linewidth,trim={0 0 0 0},clip]{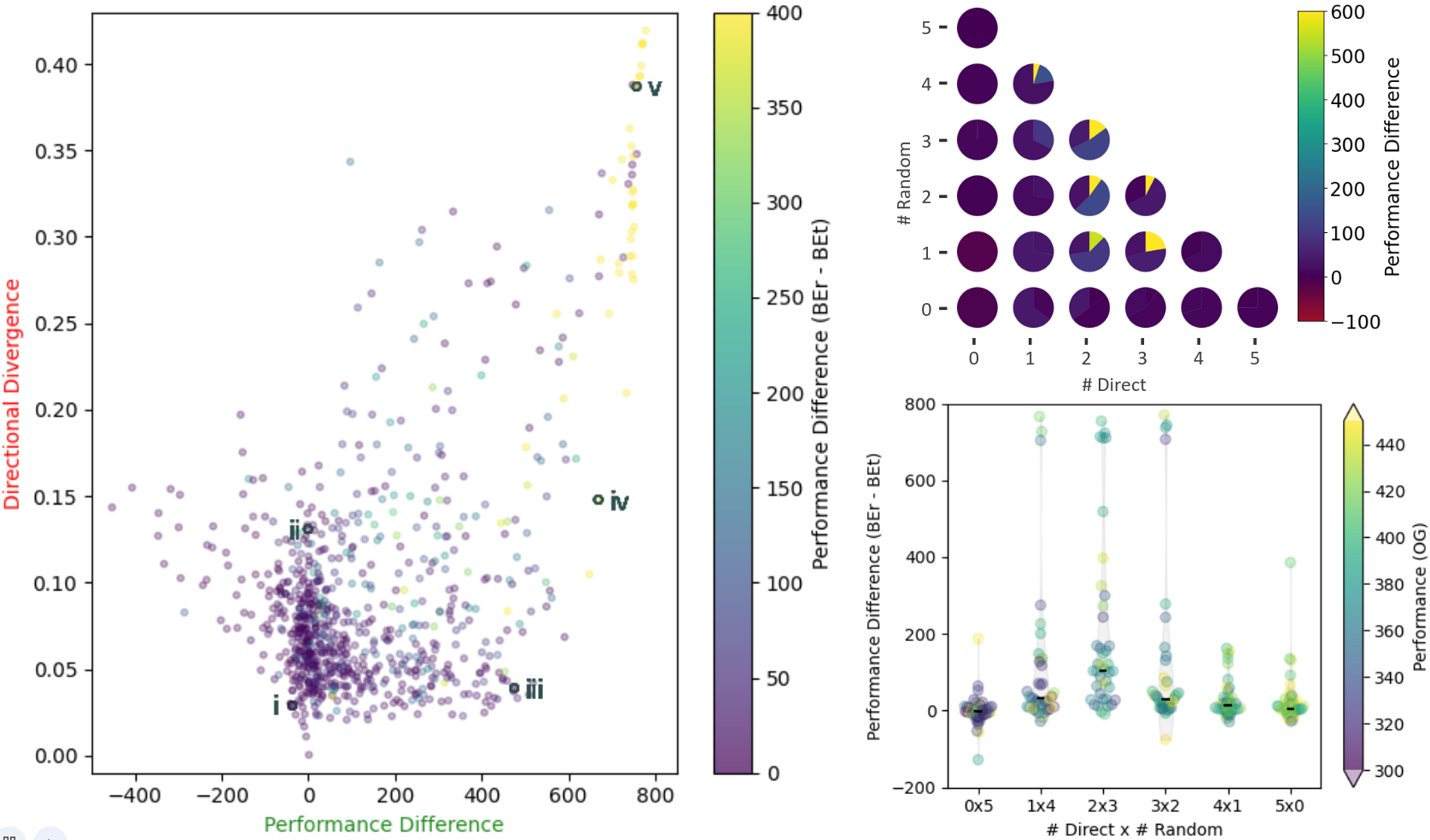}
	\end{subfigure}
	\caption{
		\textbf{Discernment for followers in mixed environments.}
		Left: scatter plot mirroring Fig. \ref{fig:dep-dirdiv}C,
		colored according to the performance difference between beacon-explorer (BEr) \& beacon-exploiter (BEt) test conditions.
		Positive values indicate asymmetric social dependence on exploiting agents, once they reached the patch, relative to exploring agents,
		i.e. discernment with respect to social status.
		Top-right: categorical stair plot mirroring Fig. \ref{fig:dep-dirdiv}F,
		showing strong discernment for follower populations in mixed direct-random environments.
		Bottom-right: plotting discernment with task performance in the training environment (time taken to reach the patch).
		}
	\label{fig:ETER}
  \end{figure}

\begin{figure}[tb]
	\centering
	\begin{subfigure}[b]{\linewidth}
	  \centering
	  \includegraphics[width=1\linewidth,trim={0 0 0 0},clip]{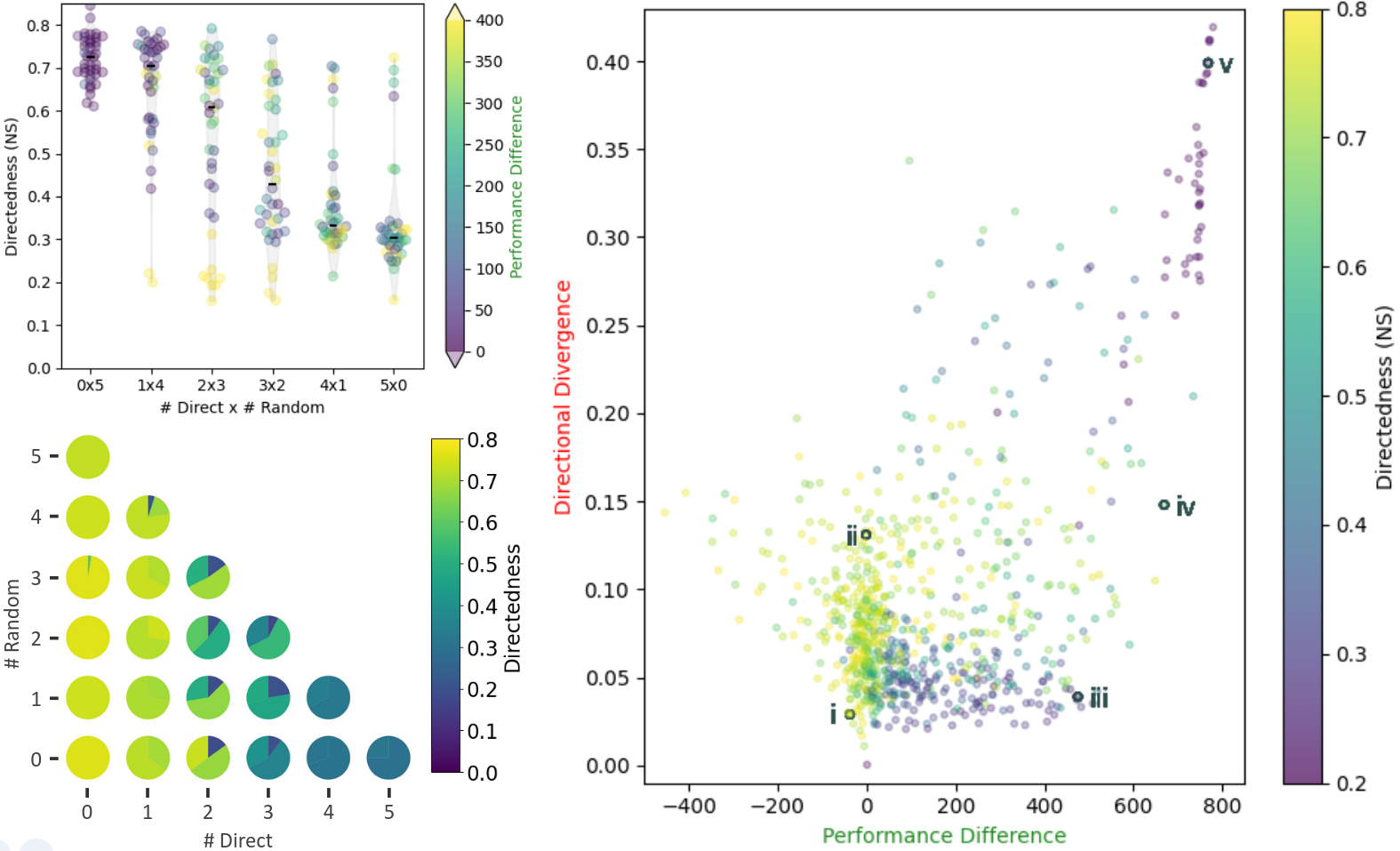}
	\end{subfigure}
	\caption{
		\textbf{Movement behavioral shift in high direct environments.}
		Directedness, a spatial correlation metric previously used to characterize individual navigation movement behavior in our model
		\cite{govoniVisuospatialNavigationDistance2024}, 
		uses information entropy to describe orientational predictability.
		Referring to Fig. 5A in \cite{govoniVisuospatialNavigationDistance2024},
		agents using straight lines and sharp turns to reach the patch corresponds to a high directedness of about 0.74,
		whereas agents using spinning or ratcheting corresponds to a low directedness from 0.3 - 0.65.
		Top-left: directedness for N=5 training environments, colored with performance difference.
		Bottom-left: directedness overlayed with the categorical stair plot of Fig. \ref{fig:dep-dirdiv}F.
		Right: scatter plot of Fig. \ref{fig:dep-dirdiv}C.
		The three plots demonstrate the followers as having the lowest directedness, 
		the hybrids in high direct environments have low directedness,
		and the individual navigators have high directedness.
		}
	\label{fig:dirent}
  \end{figure}

\begin{figure}[htb]
	\centering
	\includegraphics[width=1\linewidth,trim={0 0 0 0},clip]{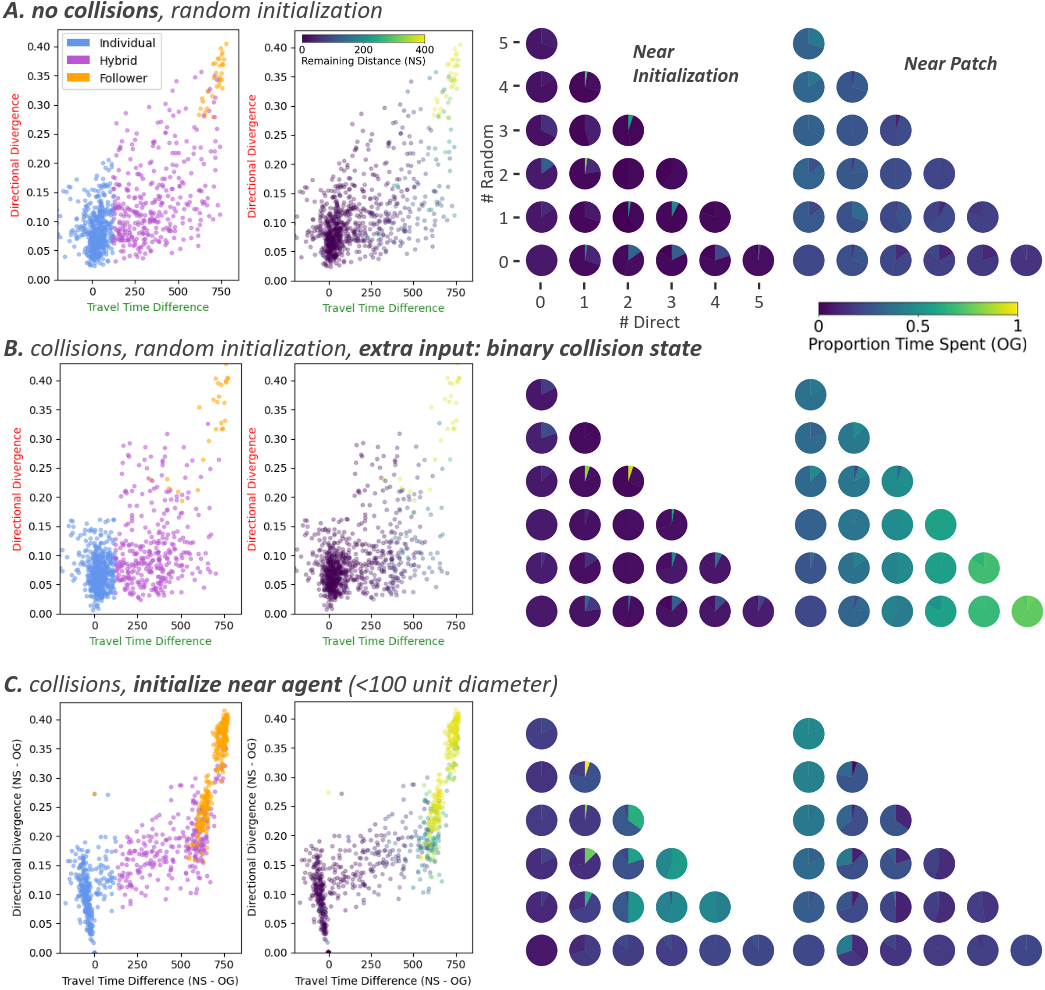}
	\caption{
		\textbf{Model Extensions: Extra.}
		Results rerun with 3 changes to simulation parameters.
		Plots follow Fig. \ref{fig:dep-dirdiv}C/\ref{fig:types}B.
		A: 
		no collisions between agents.
		B: 
		focal agent receives extra binary input indicating current collision status to the linear output layer.
		C:
		untrained agents initialize within 100 spatial units of focal agent,
		where the test perturbations is NS-OG rather than NS-BEt (Fig. \ref{fig:intro}right-bottom) 
		given the unique initialization procedure \textcolor{red}{(to elaborate later)}.
		}
	\label{fig:modelext-extra}
  \end{figure}

\begin{figure}[tb]
	\centering
	\includegraphics[width=.7\linewidth,trim={0 0 0 0},clip]{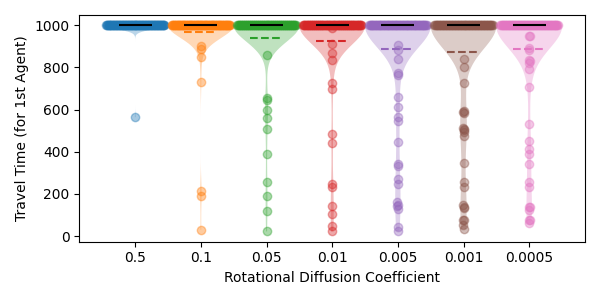}
	\includegraphics[width=.7\linewidth,trim={0 0 0 0},clip]{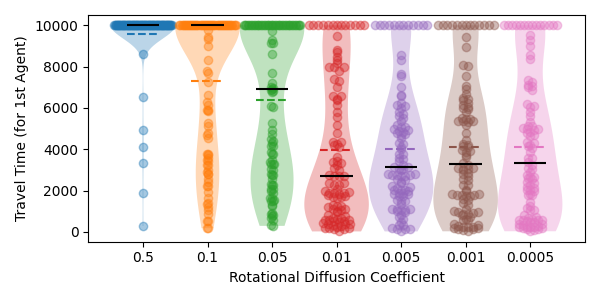}
	\includegraphics[width=.7\linewidth,trim={0 0 0 0},clip]{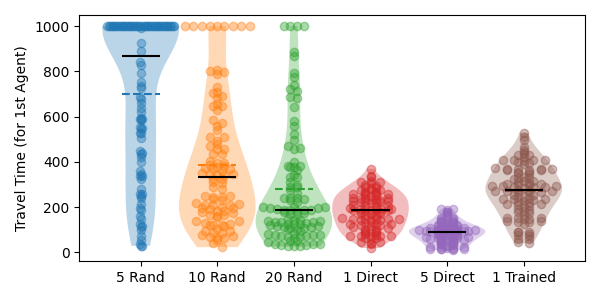}
	\caption{
		\textbf{First passage time.}
		Top: time taken for a single random agent to find the patch, with varying rotational diffusion coefficients.
		If no agent has found the patch by the end of the time limit, it is plotted at that limit.
		Middle: as above, but for 10x the limit used in this study.
		Bottom: for varying numbers of random agents (rotational diffusion coefficient = 0.01), for direct agents, and for the median agent trained in a non-social environment.
		Horizontal lines designate 100-seed population medians (black) and means (dashed colored).
		}
	\label{fig:first-passage-time}
  \end{figure}

\begin{figure}[htb]
	\centering
	\includegraphics[width=1\linewidth,trim={0 0 0 0},clip]{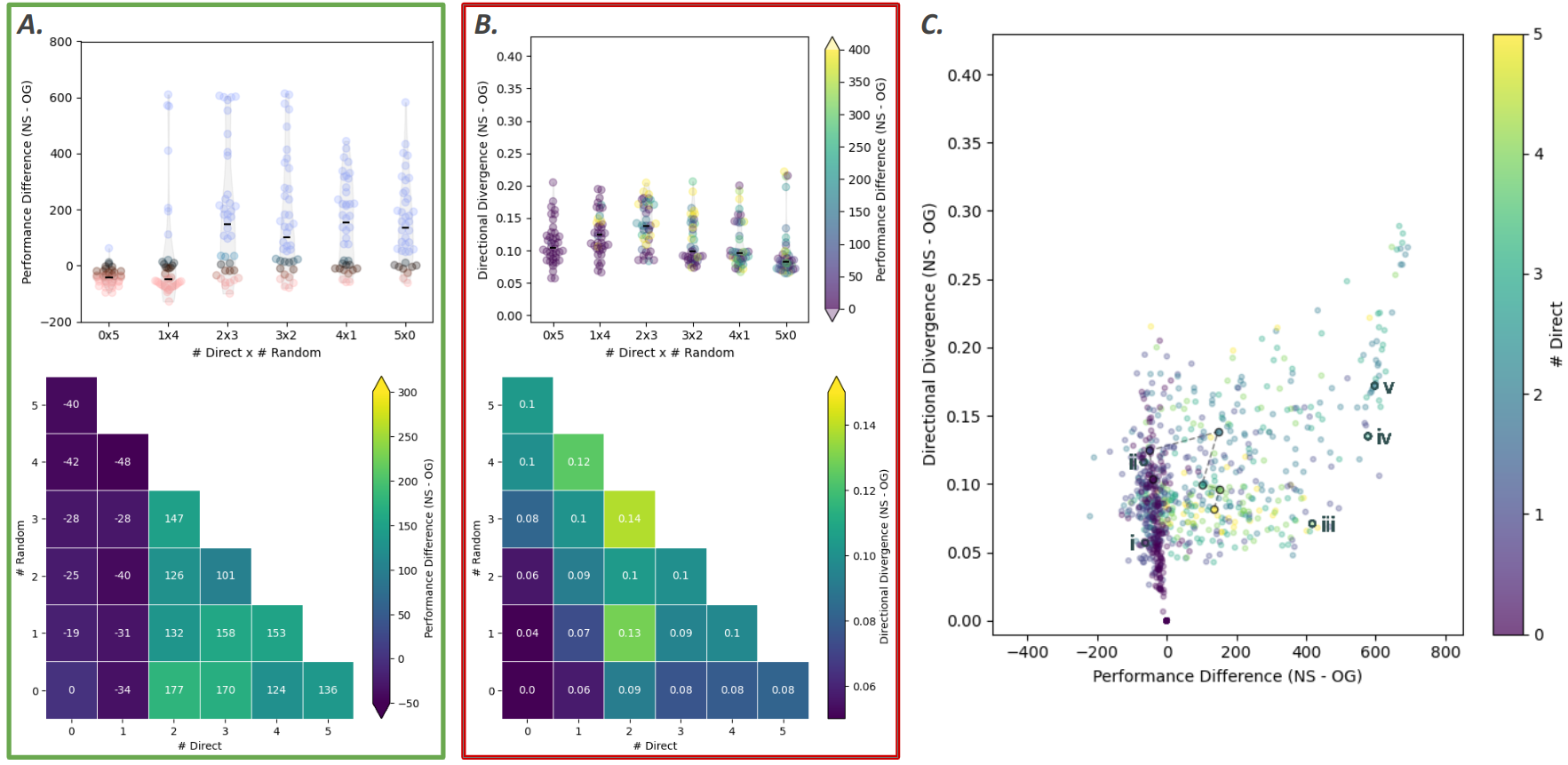} 
	\caption{
		\textbf{Learned task \& spatial dependencies - NS-OG perturbation.}
		Mirrors Fig. \ref{fig:dep-dirdiv}A-C in text,
		though here comparing between non-social (NS) test \& original (OG) training environments.
		While the performance difference trends remain similar as the NS-BEt results reported in the main text,
		the directional divergence shows a different pattern in the stair plot.
		While the NS and BEt test environments both have identical social environments for each training condition,
		the OG training environment varies social density according to condition.
		This appears to surface in the directional divergence metric,
		as more agents in the system would lead to greater overall spatial variation.
		Thus, the scatter plot above displays a similar pattern as in the main text,
		since the total number of agents are kept constant at N=5.
		}
	\label{fig:dependencies-flip}
  \end{figure}


\end{document}